%% file: main_acl.tex
\pdfoutput=1

\documentclass[11pt]{article}

\usepackage[dvipsnames]{xcolor}
\definecolor{color1}{HTML}{006EB8}
\definecolor{color2}{HTML}{009B55}
\definecolor{color3}{HTML}{00A99A}
\definecolor{color4}{HTML}{3C8031}
\definecolor{color5}{HTML}{006795}
\definecolor{color6}{HTML}{00AEB3}
\definecolor{mygray}{gray}{0.93}
\definecolor{mygreen}{HTML}{3FBC9D}
\definecolor{arsenic}{rgb}{0.23, 0.27, 0.29}

\usepackage{acl}
\usepackage[utf8]{inputenc}
\usepackage[T1]{fontenc}
\usepackage{hyperref} 
\usepackage{url}
\usepackage{booktabs}
\usepackage{amsfonts}
\usepackage{nicefrac}
\usepackage{microtype}
\usepackage{wrapfig}
\usepackage{subcaption}
\usepackage{lipsum}  
\usepackage{times}
\usepackage{tikz}
\usepackage{latexsym}
\usepackage{graphicx}
\usepackage{placeins}
\usepackage{bbm}
\usepackage{multirow}
\usepackage{enumitem}
\usepackage{tikz}
\usepackage[export]{adjustbox}
\usepackage{amsmath}
\usepackage{amssymb}
\usepackage{mathtools}
\usepackage{amsthm}
\usepackage{nccmath}
\usepackage{colortbl}
\usepackage{pifont}
\newcommand{\cmark}{\ding{51}}
\newcommand{\xmark}{\ding{55}}
\newcommand{\greencmark}{{\color{green}\cmark}}
\newcommand{\redxmark}{{\color{red}\xmark}}

\input{sections/math_commands.tex}

\newcommand{\taskcl}[1]{\textsc{CodeTask-CL}}
\newcommand{\method}[1]{}

\title{Exploring Continual Learning for Code Generation Models}

\author{Prateek Yadav$^1$\Thanks{ Work conducted during an internship at Amazon}, Qing Sun$^2$ \Thanks{ Corresponding author qinsun@amazon.com}, Hantian Ding$^2$, Xiaopeng Li$^2$, Dejiao Zhang$^2$, \\ \bf Ming Tan$^2$, Xiaofei Ma$^2$, Parminder Bhatia$^2$, Ramesh Nallapati$^2$, \\ \bf Murali Krishna Ramanathan$^2$, Mohit Bansal$^{1,3}$, Bing Xiang$^2$ \\
University of North Carolina, Chapel Hill$^1$, AWS AI Labs$^2$, Amazon Alexa AI$^3$ \\ 
\small{\{\texttt{praty, mbansal}\}\texttt{@cs.unc.edu}} \\ \small{\{\texttt{qinsun, dhantian, xiaopel, dejiaoz, mingtan, xiaofeim,}} \\ \small{\texttt{parmib, rnallapa, mkraman, mobansal, bxiang}\}\texttt{@amazon.com}}
}

\begin{document}
\maketitle

\begin{abstract}
Large-scale code generation models such as Codex and CodeT5 have achieved impressive performance. However, libraries are upgraded or deprecated very frequently and re-training large-scale language models is computationally expensive.
Therefore, Continual Learning (CL) is an important aspect that remains under-explored in the code domain. 
In this paper, we introduce a benchmark called \taskcl{} that covers a wide range of tasks, including code generation, translation, summarization, and refinement, with different input and output programming languages. 
Next, on our \taskcl{} benchmark, we compare popular CL techniques from NLP and Vision domains. 
We find that effective methods like Prompt Pooling (PP) suffer from \textit{catastrophic forgetting} due to the unstable training of the prompt selection mechanism caused by stark distribution shifts in coding tasks.
We address this issue with our proposed method, \textit{Prompt Pooling with Teacher Forcing} (PP-TF), that stabilizes training by enforcing constraints on the prompt selection mechanism and leads to a 21.54\% improvement over Prompt Pooling.
Along with the benchmark, we establish a training pipeline that can be used for CL on code models, which we believe can motivate further development of CL methods for code models. Our code is available at \href{https://github.com/amazon-science/codetask-cl-pptf}{https://github.com/amazon-science/codetask-cl-pptf}.
\end{abstract}

\input{sections/intro.tex}

\input{sections/related.tex}
\input{sections/method2.tex}

\input{sections/experiments.tex}

\section{Conclusion}
In conclusion, we have introduced a novel benchmark, \taskcl{}, tailored to cover a broad spectrum of tasks in the code domain, aiming to fuel advancements in Continual Learning (CL) for large-scale code generation models. Our study underscores the shortfalls of popular CL methods like Prompt Pooling when applied to coding tasks, predominantly due to catastrophic forgetting. However, we demonstrate that our proposed method, Prompt Pooling with Teacher Forcing (PP-TF), can effectively mitigate this issue, leading to a significant improvement of 21.54\% over the baseline. Furthermore, we establish a comprehensive training pipeline catering to CL on code models. We believe that our contributions, both in the form of the \taskcl{} benchmark and the PP-TF method, will ignite further exploration and innovation in CL techniques specifically designed for the dynamic and evolving realm of code generation.

\section*{Limitations}
This work primarily focuses on evaluating the efficacy of existing continual learning (CL) methods for code generation models. It is important to note that many of these methods were specifically designed for natural language processing or computer vision domains and may not directly transfer to the code generation domain. Nevertheless, we have made efforts to identify and address any issues encountered during our analysis. It should be acknowledged, however, that the scope of our work is limited by the selection of methods and the benchmark used. While we have utilized the most popular CL methods from various categories, there may be methods that have not been included in this study due to their inefficacy in natural language processing or computer vision tasks but may be effective in code generation. As such, we encourage further research within the community to explore the potential of CL methods for code-generation models.

\section*{Acknowledgment}
We thank Amazon for the \textit{Amazon Post-Internship Fellowship} award that supported Prateek during this work. We also thank all the reviewers for their feedback on the paper. 

\bibliography{custom}
\bibliographystyle{acl_natbib}

\appendix

\input{sections/appendix.tex}

\end{document}

%% file: sections/math_commands.tex
\usepackage{amsmath,amsfonts,bm}

\def\eqref#1{equation~\ref{#1}}

\def\1{\bm{1}}

\def\vx{{\bm{x}}}
\def\vy{{\bm{y}}}

\DeclareMathAlphabet{\mathsfit}{\encodingdefault}{\sfdefault}{m}{sl}
\SetMathAlphabet{\mathsfit}{bold}{\encodingdefault}{\sfdefault}{bx}{n}



%% file: sections/intro.tex
\section{Introduction}

Code generation models \cite{nijkamp2022codegen,wang2021codet5,lecoderl,fried2022incoder} can increase the productivity of programmers by reducing their cognitive load.
These models require significant computation to train as they have billions of parameters trained on terabytes of data. Hence, they are trained once and are then used repeatedly for several downstream applications. However, as software development constantly evolves with new packages, languages, and techniques \cite{ivers2020softevolution}, it is expensive to retrain these models.
Therefore, it is essential to continually improve these models to avoid errors, generate optimized code, and adapt to new domains and applications.

\begin{figure}[t!]
    \centering
    \includegraphics[width=\linewidth]{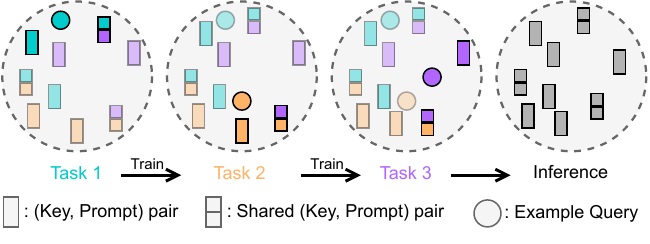}
    \captionof{figure}{\label{fig:main}We show the process of prompt selection for Prompt Pooling with Teacher Forcing when learning multiple tasks sequentially. First, we initialize the prompt pool with (key, prompt) pairs (denoted by rectangles). Next, each (key, prompt) pair is assigned to either a single task or is shared by two tasks (denoted by colors). When learning Task 1 (green color), we obtain the query (green circle) for a given example and select the top-k (k=2 here) pairs from the assigned (key, prompt) pairs, highlighted in the figure. 
    These selected pairs are then trained for the example. A similar process is followed for subsequent tasks. During inference, we remove task assignments and select the top-k pairs across all the pairs.
    }
\end{figure}

We explore continual learning (CL) \cite{ring1998child,thrun1998lifelong} abilities of code-generation models and aim to improve them. Specifically, we present a \taskcl{} benchmark for code-based CL and aim to train a model on sequentially presented tasks with different data distributions without suffering from catastrophic forgetting (CF) \cite{mccloskey1989catastrophic}. This occurs when the model overfits the current task, resulting in a decline in performance on previously learned tasks.

Given the lack of CL benchmarks for the code domain, we create a benchmark called \taskcl{} using existing datasets. It consists of tasks like code completion \cite{iyer2018mapping,iyer2019learning,clement2020pymt5}, code translation \cite{chen2018tree,lachaux2020unsupervised}, code summarization \cite{wang2020trans,Wang2020CoCoGUMCC}, and code refinement \cite{tufano2019empirical}. This benchmark presents a new and challenging scenario as it necessitates the adaptation of the model to varying input and output programming languages. Along with this benchmark, we also present a training framework to easily apply CL methods to code generation models. 

Next, we evaluate the effectiveness of popular CL methods from NLP and Vision domains in the context of code generation models. We consider prompting methods \cite{wang2022learning,li2021prefixtuning} and experience-replay \cite{de2019continual} due to their good performance for pre-trained models \cite{wu2022pretrainedcl}.
We also experiment with Prompt Pooling (PP) \cite{wang2022l2p}, an effective prompting-based method for CL in the vision domain.
Our results show that Prompt Pooling suffers from catastrophic forgetting on our proposed \taskcl{} benchmark because of the complex distribution shift from varying input and output programming languages across tasks. With further investigation, we find that the unconstrained prompt selection mechanism leads to an unstable training problem. To address this, we propose our method \textit{Prompt Pooling with Teacher Forcing} (\textsc{PP-TF}), which imposes constraints on prompt selection during training by assigning certain prompts to fixed tasks during training (see Figure \ref{fig:main}). This results in stable training and better performance.
Interestingly, we find when a replay buffer is available, the simple experience-replay \cite{de2019continual} method outperforms other CL methods and achieves performance similar to a multitask baseline \cite{Crawshaw2020MultiTaskLW} where all tasks are provided at once.

In summary, our contributions include: (1) being the first study on CL for code generation tasks, (2) establishing a benchmark and a novel pipeline that supports CL for code generation to motivate future work, (3) identifying and addressing the unstable training issue of Prompt Pooling through our proposed method PP-TF, and (4) discussion on the best CL methods to use in different use cases.

%% file: sections/related.tex
\section{Related Work}
\paragraph{Code Generation Models.} Code generation and language modeling for source code is an emerging research field experiencing active growth. Several model architectures have been examined recently, including encoder-only models \cite{feng2020codebert,guo2020graphcodebert}, encoder-decoder models \cite{ahmad-etal-2021-unified,wang2021codet5}, and decoder-only models \cite{nijkamp2022codegen,chen2021evaluating,Nijkamp2022ACP}. However, none of these models have been studied in the context of continual learning.

\paragraph{Continual Learning.} There are various methods for Continual Learning (CL) and they fall into three categories: \textit{Regularization}, \textit{Replay}, and \textit{parameter isolation} methods. \textbf{Regularization methods} \cite{Kirkpatrick2018overcoming,zenke2017continual,schwarz2018progress} assign importance to model components and add regularization terms to the loss function. \textbf{Replay methods} \cite{de2019continual,rebuffi2017icarl,lopez2017gradient,chaudhry2018efficient} retain a small memory buffer of data samples and retrain them later to avoid catastrophic forgetting (CF). 
\textbf{Parameter isolation methods}, such as prompting-based methods \cite{wang2022learning,wang2022dualprompt,li2021prefixtuning,liu2021gpt,qin-eisner-2021-learning}, introduce or isolate network parameters for different tasks. For a more comprehensive overview of all CL methods, we refer the reader to \citet{delange2021clsurvey,biesialska-etal-2020-continualsurvery}.

To the best of our knowledge, there are currently no studies or benchmarks for CL on code generation models. Therefore, we evaluate the effectiveness of prompting \cite{wang2022learning,li2021prefixtuning} and experience replay \cite{chaudhry2018efficient,buzzega2020der} based methods, which have demonstrated strong performance in CL on large pretrained models \cite{raffel2019exploring}. We do not consider regularization methods as they are not effective in continually learning large-scale pretrained models \cite{wu2022pretrained}.
Next, we discuss our proposed benchmark and methods.

%% file: sections/method2.tex
\section{\taskcl{} Benchmark}
\label{sec:benchmarks}

We present the \taskcl{} benchmark to assess the CL abilities of code generation models. 
We also provide a novel training pipeline that can be used to continually train and evaluate code generation models.
All of the datasets used to create the \taskcl{} benchmark are available under the MIT license and more details on the dataset splits and input-output domains are in Table \ref{tab:app_dataset}.

\subsection{Coding Tasks}

\noindent \textbf{Code Generation} aims to generate a code snippet from a natural language description. We use the CONCODE dataset \cite{iyer2018mapping} which is a collection of tuples that consist of natural language descriptions, code environments, and code snippets, obtained from approximately 33,000 Java projects on GitHub. 
The objective of the study is to generate class member functions utilizing the natural language descriptions and class environment.

\noindent \textbf{Code Summarization} aims to generate a summary for a piece of code. We use the CodeSearchNet dataset \cite{husain2019codesearchnet}, which consists of six programming languages (Python, Java, JavaScript, PHP, Ruby, and Go). The data for this task consists of the first paragraph of each documentation.

\noindent \textbf{Code translation} refers to the transformation of a program written in a particular programming language into another language while maintaining its functionality. We use the Java $\rightarrow$ C\# dataset compiled by \citet{lu2021codexglue} that provides pairs of code that perform the same tasks.

\noindent \textbf{Code Refinement} aims to improve the code by fixing bugs within the code automatically. We use the dataset provided by \citet{tufano2019empirical} consisting of pairs of faulty and corrected Java functions.

\subsection{Evaluation}
\label{sec:metrics}
Next, we define the  metrics used to evaluate a model continually on these datasets. We follow \citet{lu2021codexglue} and evaluate each task using BLEU \cite{papineni2002bleu}. 
We follow \cite{chaudhry2018efficient} to continually evaluate model's performance. We measure the \textit{average BLEU} after learning all the tasks as, $\text{<BLEU>} = \frac{1}{N} \sum_{k=1}^{N} b_{N,k}$, where $N$ is the total number of tasks and $b_{i,j}$ represents the BLEU score on task $j$ after learning task $i$. Additionally, we report the average forgetting metric, denoted by $\text{<Forget>}$, to assess the model's ability to retain performance on previously learned tasks. This metric is calculated as the average difference between the maximum accuracy obtained for each task $t$ and its final accuracy, given by $\text{<Forget>} = \frac{1}{N-1} \sum_{t=1}^{N-1} (\max_{k \in {1, \hdots, N-1}} b_{k,t} - b_{N,t})$.

\begin{table*}[tbh]
\centering
\Huge
\resizebox{0.99\linewidth}{!}{  

\begin{tabular}{cc|cccc|ccc}
\toprule
\textbf{Method ($\downarrow$)} & \textbf{Replay [5k]} & \textbf{Code Gen.} & \textbf{Code Trans.} & \textbf{Code Summ.} & \textbf{Code Ref.} & \textbf{<BLEU\textsubscript{Test}>} & \textbf{<BLEU\textsubscript{Val}>} & \textbf{<Forget\textsubscript{Val}>} \\
\midrule

\rowcolor{mygray}
\textbf{Sequential FT} & \redxmark & 6.42 & 2.76 & 3.13 & 77.75 & 22.52 & 22.44 & 39.64 \\
\rowcolor{mygray}
\textbf{MTL} & \redxmark & 32.24 & 74.87 & 14.69 & 79.23 & 50.26 & 49.25 & - \\
\rowcolor{mygray}
\textbf{Individual FT} &  \redxmark & 38.61 & 83.34 & 14.32 & 77.73 & 53.50 & 52.68 & - \\
\midrule

\textbf{Shared Prompts} & \redxmark & 0.63 & 6.75 & 0.37 & 78.5 & 21.56 & 21.71 & 30.33 \\
\textbf{Shared Prompts + ER} & \greencmark & 13.82 & 45.87 & 14.36 & 78.64 & 38.17 & 36.93 & 8.46 \\
\textbf{Task Specific Prompts} & \redxmark & 22.93 & 65.37 & 14.57 & 78.81 & \textbf{45.42} & \textbf{44.56} & 0.00 \\
\midrule
\textbf{Prompt Pooling (PP)} & \redxmark & 2.41 & 7.47 & 2.62 & 78.67 & 22.79 & 23.10 & 27.43 \\
\textbf{Prompt Pooling (PP) + ER} & \greencmark & 16.33 & 50.96 & 13.13 & 78.71 & 39.78 & 38.47 & 6.41 \\
\textbf{PP + Teacher Forcing} & \redxmark & 24.28 & 59.37 & 14.15 & 79.50 & \textbf{44.33} & \textbf{43.10} & 1.68 \\
\midrule
\textbf{CodeT5 + ER} & \greencmark & 32.92 & 77.94   & 11.74 & 78.43 & \textbf{50.26} & \textbf{49.03} & 2.22 \\
\bottomrule

\end{tabular}
}
\caption{\label{tab:task_cl} BLEU scores on the test set for the individual tasks and average BLEU ($\uparrow$) and Forgetting ($\downarrow$) metrics after sequentially learning Code Generation $\rightarrow$ Code Translation $\rightarrow$ Code summarization $\rightarrow$ Code Refinement Tasks. 
}

\end{table*}

\section{Prompt Pooling With Teacher Forcing}
\label{sec:pptf}

Prompt Pooling \cite{wang2022l2p} is a highly effective technique that possesses two key benefits. Firstly, the number of prompts required does not increase linearly with the number of tasks. Secondly, the prompts within the pool can be utilized across multiple tasks, thereby enabling the reuse of previously acquired knowledge. These abilities are advantageous in real-world scenarios, particularly when a model needs to be continually adjusted to accommodate a large number of users/tasks.

In Prompt Pooling (PP), a set of learnable prompts $P = \{P_i\}_{i=1}^M$ are defined and shared by multiple tasks.
We follow \newcite{wang2022l2p} and utilize a query and key-matching process to select the prompts for each task. This process has four steps: (1) a learnable key, represented as $k_i \in \mathbb{R}^{d}$, is defined for each prompt, resulting in a prompt pool of the form $\{(k_i, P_i)\}_{i=1}^M$; (2) a query function $q(\vx)$ is defined, which takes an input $\vx$ from a given task and produces a query vector $q_{\vx} \in \mathbb{R}^{d}$; (3) the top-$k$ keys are selected based on the cosine similarity between the query $q_{\vx}$ and all the key vectors $\{k_i\}_{i=1}^M$; (4) we obtain the final input vector $\vx_p$ by pre-pending the example $\vx$ with the prompts corresponding to the selected keys. Then $\vx_p$ is fed into the pre-trained model $f$ and we minimize the following loss function to \textit{only} optimize the selected prompts and the corresponding keys while keeping the pre-trained model fixed.

\begin{equation}
    \vspace{-2pt}
    \footnotesize
    \mathcal{L} = \mathcal{L}_{LM}(x_p, y) + \lambda \sum_{k_{s_i} \in K_s} sim(q(x), k_{s_i})
    \vspace{-3pt}
    \label{eqn:loss}
\end{equation}
where $\mathcal{L}_{LM}$ is the language modeling loss, $\vy$ is the target sequence given the input $\vx$, $K_s$ is the set of selected keys from Step (3) above.

The query-key mechanism described above is an Expectation-Maximization (EM) \cite{em1996} procedure. Given an example, we first select the top-$k$ keys based on the cosine similarity (E-Step) and then train these selected keys to pull them closer to the query (M-Step). The training is stable when all tasks are jointly learned. However, in the CL context, tasks are sequentially trained which makes training unstable. Hence, we propose \textit{Prompt Pooling with Teacher Forcing} (PP-TF) that removes the E-Step by assigning each $\{(k_i, P_i)\}$ pair to fixed tasks and only performs the M-Step of optimizing the keys. To encourage knowledge sharing, we allow a few $\{(k_i, P_i)\}$ pairs to be shared across tasks (see Figure \ref{fig:main}). With these assignments/constraints in place, when training on task $t$, we use teacher forcing to select top-$k$ prompts that are assigned to the task. Thus, for learning task $t$, our loss function becomes,

\begin{equation}
    \footnotesize
    \vspace{-2pt}
    \mathcal{L} = \mathcal{L}_{LM}(x_p, y) + \lambda \sum_{k_{s_i} \in K_s \cap K_t} sim(q(x), k_{s_i})
    \vspace{-2pt}
    \label{eqn:pptf}
\end{equation}
where, $K_t$ denotes the prompts assigned to task $t$ for teacher forcing.
As training progresses, the queries and keys learn to align in a stable manner, while also allowing for information sharing among tasks through the shared prompts. During inference, we discard the assignment for (key, prompt) pair and use cosine similarity to select the top-$k$ pairs across the whole pool.

%% file: sections/experiments.tex
\begin{figure*}[t!]
    \vspace{10pt}
  \begin{subfigure}[b]{0.19\textwidth}
    \includegraphics[width=\textwidth]{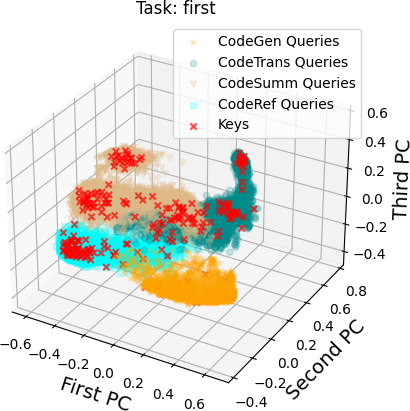}
    \caption{At Initialization}
  \end{subfigure}
  \hfill
  \begin{subfigure}[b]{0.19\textwidth}
    \includegraphics[width=\textwidth]{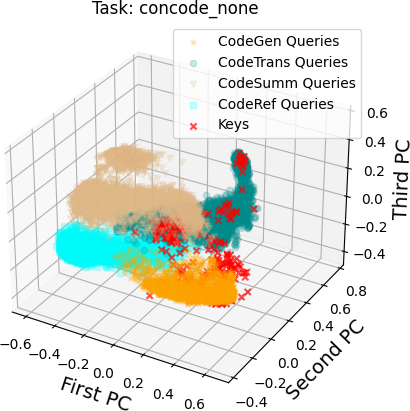}
    \caption{After CodeGen}
  \end{subfigure}
  \hfill
  \begin{subfigure}[b]{0.19\textwidth}
    \includegraphics[width=\textwidth]{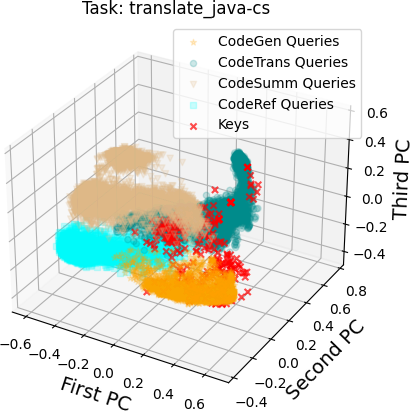}
    \caption{After CodeTrans}
  \end{subfigure}
  \hfill
  \begin{subfigure}[b]{0.19\textwidth}
    \includegraphics[width=\textwidth]{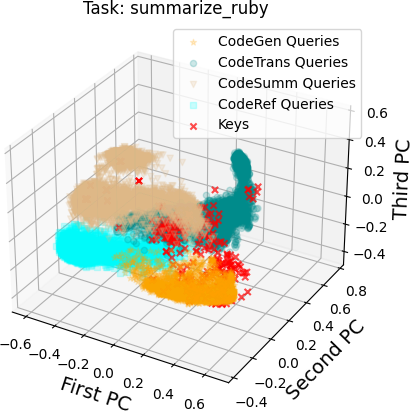}
    \caption{After CodeSumm}
  \end{subfigure}
  \hfill
  \begin{subfigure}[b]{0.19\textwidth}
    \includegraphics[width=\textwidth]{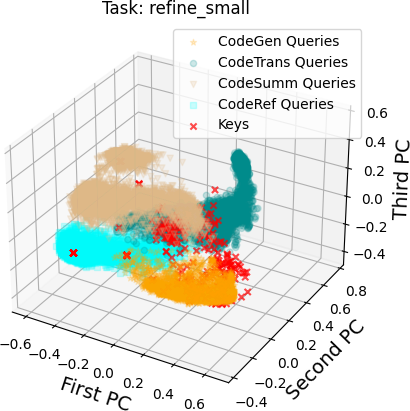}
    \caption{After CodeRef}
  \end{subfigure}
  \vspace{-7pt}
  \caption{\label{fig:evolution} We plot the evolution of keys during the training process along with the fixed queries when sequentially learning, Code Generation $\rightarrow$ Code Translation $\rightarrow$ Code summarization $\rightarrow$ Code Refinement Tasks.}
\end{figure*}

\section{Experiments}
\label{sec:experiments}

We focus on the scenario of known task identities for continual learning. This is commonly the case in code-related domains and task identities can also be determined through input and output analysis in certain situations.
In the field of NLP and Vision, methods utilizing experience replay and prompting have been highly effective for CL on large pre-trained models \cite{wang2022l2p,wang2021transprompt,wu2022pretrainedcl}. Moreover, regularization methods are shown to not work well in conjunction with pre-trained models \cite{wu2022pretrainedcl}, and hence, we skip them from our study. Next, we present these methods along with some baseline methods.

\subsection{Baselines}

\noindent \textbf{Sequential Finetuning} \cite{yogatama2019learning} updates all model parameters for every incoming task in a sequential manner. This approach has been shown to suffer from catastrophic forgetting and serves as a lower bound for CL methods.

\noindent \textbf{Individual Models} \cite{howard2018finetune} finetune a separate models for each new task. This is considered an upper bound for CL methods.

\noindent \textbf{Multitask Learning} \citep{Crawshaw2020MultiTaskLW} simultaneously learns multiple tasks at once, without experiencing distribution shift, resulting in a strong performance. For multitask learning, we prepend the task descriptors to the input and follow \citet{wang2021codet5} to ensure balanced sampling across tasks with varying dataset sizes.

\noindent \textbf{Shared Prompt Tuning (SP)} defines $M$ soft continuous prompts \cite{li-liang-2021-prefix} which are added and fine-tuned for each example from all tasks. They are trained via gradient descent while keeping the pretrained model's parameters fixed.

\noindent \textbf{Task Specific Prompt Tuning (TSPT)} defines a total of $M$ soft continuous prompts \cite{li-liang-2021-prefix} that are divided across $N$ tasks, resulting in $\lfloor\frac{M}{N}\rfloor$ task-specific prompts.

\noindent \textbf{Experience Replay (ER)} \cite{riemer2018learning} involves maintaining a memory buffer $B$ of examples from the previous task. The buffer randomly stores an equal number of samples from each past task and is used to retrain the model at later stages. Moreover, as several of the other methods outlined in this study can benefit from ER, we also include results with and without the utilization of ER.

\subsection{Main Results}
\label{sec:results}

\subsubsection{Task-CL Experiments} We use CodeT5 model \cite{wang2021codet5} as our pre-trained model when learning the \taskcl{} benchmark. In Table \ref{tab:task_cl}, we report results for a single run on the methods described above and their ER variants. For more implementation details and hyperparameters used please refer to Appendix \ref{sec:app_hyper}. 
First, we find that the popular prompt pooling demonstrates catastrophic forgetting with a test BLEU score of 22.79\%. Even when using ER with PP the performance is 39.78\% which is still much worse than other methods. In contrast, \textit{PP + TF} even without ER outperforms \textit{PP} and \textit{PP + ER} by 21.54\% and 4.55\% respectively.
Moreover, our results show that the \textit{CodeT5 + ER} method which finetunes the full CodeT5 model with ER performs the best with an average test BLEU score of 49.21\%. Please refer to Appendix \ref{sec:app_replay} for experiments on the effect of buffer size on performance.\\

\noindent \textbf{Discussion:} We find that task-specific prompts are more effective than other prompting-based CL methods. However, due to their high storage requirements that scales linearly with the number of tasks, this approach is not feasible for large-scale applications where the model needs to be adapted for a large number of users or tasks.
In contrast, a memory buffer might be available due to privacy concerns \cite{yoon2021federatedcl} in many situations. In such cases, the \textit{PP-TF} is the recommended method. Given these findings, we believe that the current Prompt Pooling based methods can be further  improved in order to reuse knowledge across tasks.

\subsubsection{Training Instability of Prompt Pooling}
To show the root of catastrophic forgetting in prompt pooling, we evaluate how queries and keys align in the representation space after learning each task. To do so, we first select a subset of 5k training samples from four tasks resulting in 20k examples. We utilize a fixed codeT5 encoder as our query function that encodes provided examples to obtain queries. These queries remain unchanged during training and the keys are initialized using the data. We then use principal component analysis (PCA) \cite{pca1901} on the queries and keys to obtain the first three principal components and plot them. After learning each task, we repeat the PCA step on the fixed queries and the updated prompt keys. 

From Figure \ref{fig:evolution}, we observe before the training starts, the keys (represented by red crosses) are evenly distributed among the queries of different tasks. However, after completing the training on the first task (CodeGen), most of the keys move toward the queries associated with that CodeGen (denoted by orange stars). This indicates that the prompts corresponding to these keys were primarily used for the CodeGen task and were trained by it. As a large portion of the prompts from the pool are utilized during the training of the CodeGen task, there are no key vectors available for allocation to the second task (CodeTrans). 
As a result, when learning the CodeTrans, some keys used for the previous task are pulled toward CodeTrans's queries and the corresponding prompts are updated. 
As each subsequent task is introduced, the key vectors are dynamically adjusted to align with the current task's queries, leading to a unstable process of matching in which updates to the key-prompt pairs are frequently in conflict with the previous tasks. Hence leading to catastrophic forgetting on the previous tasks.

%% file: sections/appendix.tex
\begin{table*}[t!]
\centering
\huge
\resizebox{0.8\linewidth}{!}{  

\begin{tabular}{ccc|cc|ccc}
\toprule
\textbf{Scenario} & \textbf{Task} & \textbf{Dataset Name} & \textbf{Input} & \textbf{Output} & \textbf{Train} & \textbf{Validation} & \textbf{Test} \\
\midrule

\multirow{4}{*}{\textbf{Task-CL}} & Generation & CONCODE & English & Java & 100k & 2k & 2k \\
& Translation & CodeTrans & Java & C\# & 10k & 0.5k & 1k \\
& Sumarization & CodeSearchNet & Ruby & English & 25k & 1.4k & 1.2k \\
& Refinement & BFP & Java & Java & 46k & 5.8k & 5.8k \\
\bottomrule

\end{tabular}
}

\caption{\label{tab:app_dataset} Table providing the Dataset Statistics for the task used in \taskcl{} benchmark. We specify the input and output domains along with the split sizes for train, validation, and test sets.
}
\end{table*}

\begin{table*}[tbh]
\centering
\huge
\resizebox{0.99\linewidth}{!}{  

\begin{tabular}{cc|cccc|ccc}
\toprule
\textbf{Method ($\downarrow$)} & \textbf{Buffer Size} & \textbf{Code Gen.} & \textbf{Code Trans.} & \textbf{Code Summ.} & \textbf{Code Ref.} & \textbf{<BLEU\textsubscript{Test}>} & \textbf{<BLEU\textsubscript{Val}>} & \textbf{<Forget\textsubscript{Val}>} \\
\midrule

\multirow{5}{*}{\textbf{CodeT5 + ER}} & 100 & 24.11 & 61.87 & 10.72 & 77.82 & 43.63 & 41.25 & 14.18 \\
 & 500  & 29.39 & 57.56   & 11.33 & 78.70 & 44.25 & 40.1 & 11.42 \\
& 1000 & 28.23 & 73.33   & 12.06 & 78.03 & 47.91 & 46.74 & 6.98 \\
& 2000 & 31.10  & 75.52   & 11.85 & 77.58 & 49.01 & 47.59 & 5.99 \\
& 5000 & 32.92 & 77.94   & 11.74 & 78.43 & \textbf{50.26} & \textbf{49.03} & \textbf{2.22} \\
\midrule
\textbf{MTL} & - & 32.24 & 74.87 & 14.69 & 79.23 & 50.26 & 49.25 & - \\
\rowcolor{mygray}
\textbf{Individual FT} & - & 38.61 & 83.34 & 14.32 & 77.73 & 53.50 & 52.68 & - \\
\bottomrule

\end{tabular}
}

\caption{\label{tab:app_replay} Table showing performance on each task as we vary the Buffer Size when sequentially learning Code Generation $\rightarrow$ Code Translation $\rightarrow$ Code summarization $\rightarrow$ code Refinement Tasks.
}

\end{table*}

\section{Appendix}
\label{sec:appendix}

\subsection{Implementation Details}
\label{sec:app_hyper}

In our experiments, we report the results of a single run. We used the \textit{CodeT5-small} model \cite{wang2021codet5} with 60M parameters from Huggingface \cite{wolf2019huggingface}, which is an encoder-decoder model pre-trained on CodeSearchNet \cite{husain2019codesearchnet}. We use a separate and fixed codeT5 encoder model as the query function to encode the input examples for prompt pooling. For all prompting-related experiments, the CodeT5 model remains frozen and only the prompts are finetuned. In cases where we have ER with prompting methods, the ER is also applied while finetuning the prompts. Our prompt pool consisted of 500 prompts, with 100 prompts being selected to prepend to examples for each task. For the Shared Prompts method, we utilized 100 prompts that are used for all the tasks. For the Task-Specific Prompt method, we utilized different 100 prompts for each task. Unless otherwise specified, we used a buffer size of 5000 examples for all methods employing ER. The Adam \cite{kingma2014adam} optimizer was utilized, along with early stopping. The hyperparameters for our experiments were taken from \citet{wang2021codet5}, and the tasks from \taskcl{} benchmark were learned in random order specified in Table \ref{tab:task_cl}. 
The results of our experiments included the Average validation and test BLEU scores, as well as the forgetting metric on the validation set. The implemntation of BLEU was taken from the CodeT5 paper \cite{wang2021codet5}. We ran experiments on a single A6000 GPU with 48 GB of memory with total computation of 14 GPU days. 

\subsection{Data Statistics for \taskcl{} Benchmark}
Table \ref{tab:app_dataset} shows the train, validation, and test data sizes for all the tasks used in the \taskcl{} benchmark. We also present the input and output domains for each of the individual tasks. Given the input and output domains for these tasks are starkly different this makes this benchmark challenging as the distribution shift is large. Please refer to Section \ref{sec:benchmarks} in the main paper for more details about the benchmark. All of the datasets used to create the \taskcl{} benchmark are available under the MIT license.

\subsection{Impact of Buffer Size on ER Performance.}
\label{sec:app_replay}
If ER replay is possible, we find that \textit{CodeT5 + ER} is the most performant method. We go on to further assess the impact of buffer size on the performance. In Table \ref{tab:app_replay}, we present the aggregated results for a total buffer size of 100, 500, 1000, 2000, and 5000.
Our findings suggest that the is an increase in performance as the buffer size increases. We observe that CodeT5 + ER with a small buffer size of 100 examples outperforms PP + ER (5k examples) by 3.85\% respectively. Moreover, CodeT5 + ER with a buffer size of 1000 outperforms the best method without ER.
Our findings are in line with that of \citet{scialom2022continualt0} and demonstrate that whenever possible, we should use ER with pretrained models. Although in cases with no buffer with a large number of tasks, \textit{PP + TF} is the best method to use.